\documentclass[runningheads]{llncs}

\usepackage{eccv}

\usepackage{eccvabbrv}

\usepackage{graphicx}
\usepackage{booktabs}
\usepackage{amsmath}
\usepackage{amssymb}
\usepackage{bm}
\usepackage{colortbl}
\usepackage{enumitem}
\usepackage{graphicx}
\usepackage{makecell}
\usepackage{multirow}
\usepackage{soul}
\usepackage{tabularx}
\usepackage{wrapfig}
\usepackage{xcolor}

\newcommand{\swapslot}[1]{{\textbf{#1}}}
\newcommand{\ourresult}[1]{{\cellcolor{gray!20}{#1}}}

\newcommand{\template}[1]{{\small{\textquotedblleft
#1\textquotedblright}}}
\newcommand{\templatee}[1]{{\small\textquotedblleft
#1\textquotedblright}}
\newcommand{\first}[1]{\textbf{\textcolor{blue}{#1}}}
\newcommand{\second}[1]{\textbf{\textcolor{black}{#1}}}
\newcommand{\gray}[1]{\textcolor{lightgray}{#1}}

\usepackage{hyperref}

\newcommand{\ours}{ArtVLM}
\usepackage{orcidlink}

\begin{document}

\title{\ours{}: \textbf{A}ttribute \textbf{R}ecognition \textbf{T}hrough \\\textbf{V}ision-Based Prefix \textbf{L}anguage \textbf{M}odeling} 

\titlerunning{ArtVLM: Attribute Recognition Through Vision-Based PrefixLM}

\author{William Yicheng Zhu\inst{1}*\orcidlink{0009-0006-7899-0288} \and
Keren Ye\inst{1}*\orcidlink{0000-0002-7349-7762} \and
Junjie Ke\inst{1}\orcidlink{0009-0000-5846-9002} \and
Jiahui Yu\inst{2}†\orcidlink{0000-0002-7085-834X} \and \\
Leonidas Guibas\inst{1}\orcidlink{0000-0002-8315-4886} \and
Peyman Milanfar\inst{1}\orcidlink{0000-0003-1455-7662} \and
Feng Yang\inst{1}\orcidlink{0000-0001-6195-2089}}

\authorrunning{W. Y. Zhu et al.}
\institute{$^1$ Google Research   $^2$ OpenAI
}

\maketitle
\let\thefootnote\relax\footnote{\scriptsize{* Equal contribution. \newline † Work done at Google.}}

\begin{abstract}
    Recognizing and disentangling visual attributes from objects is a foundation to many computer vision applications. While large vision-language representations like CLIP had largely resolved the task of zero-shot object recognition, zero-shot visual attribute recognition remains a challenge because CLIP's contrastively-learned vision-language representation cannot effectively capture object-attribute dependencies. In this paper, we target this weakness and propose a sentence generation-based retrieval formulation for attribute recognition that is novel in 1) explicitly modeling a to-be-measured and retrieved object-attribute relation as a conditional probability graph, which converts the recognition problem into a dependency-sensitive language-modeling problem, and 2) applying a large pretrained Vision-Language Model (VLM) on this reformulation and naturally distilling its knowledge of image-object-attribute relations to use towards attribute recognition. Specifically, for each attribute to be recognized on an image, we measure the visual-conditioned probability of generating a short sentence encoding the attribute's relation to objects on the image. Unlike contrastive retrieval, which measures likelihood by globally aligning elements of the sentence to the image, generative retrieval is sensitive to the order and dependency of objects and attributes in the sentence. We demonstrate through experiments that generative retrieval consistently outperforms contrastive retrieval on two visual reasoning datasets, Visual Attribute in the Wild (VAW), and our newly-proposed Visual Genome Attribute Ranking (\href{https://github.com/google-research/google-research/tree/master/attribute_with_prefixlm}{VGARank}).
\end{abstract}

\section{Introduction}
\label{sec:intro}

Understanding attributes associated with objects in an image is essential for many computer vision applications, including content recommendation, visual reasoning, and text-to-image generative models. While supervised learning techniques such as classification~\cite{Simonyan2015very,krizhevsky2017imagenet,He_2016_CVPR}, detection~\cite{ren2015faster,redmon2016you,He_2017_ICCV}, and segmentation models~\cite{ronneberger2015u,chen2017deeplab} have made significant progress in object recognition tasks, directly adding a branch for object-agnositic attribute prediction~\cite{farhadi2009describing,ferrari2007learning,patterson2016coco} can result in incorrect and counterfactual outputs since it fails to model the co-dependency between attributes and the objects. Other existing attribute learning methods rely heavily on human annotations \cite{Zhang_2021_CVPR,Anderson_2018_CVPR,lampert2009learning,jayaraman2014zero,al2016recovering} to address this dependency, but this makes them expensive and hard to scale. All things considered, how to properly establish object-attribute relationship at scale remains an open problem.

Large-scale image-text foundation models such as CLIP~\cite{radford2021learning} and ALIGN~\cite{jia2021scaling} inspired us to explore their potential for attribute learning. These models have learned from a vast amount of noisy image-text pairs from the web, adequately utilizing self-supervised learning to benefit from easily accessible data sources. They have shown exceptional performance in zero-shot object recognition~\cite{Rao_2022_CVPR,yao2021cpt,Zhong_2022_CVPR,Wang_2022_CVPR,Ma_2022_CVPR,Shi_2022_CVPR,Materzynska_2022_CVPR,Pham_2022_ECCV} through image-text similarity measurement, a method which we refer to as ``contrastive retrieval''.

\begin{figure}[t]
    \centering
    \caption{Prefix language modeling and generative prompting. During pretraining, the image-conditioned prefix language model (prefixLM) learns to generate the captions associated with images, and through this way it curates knowledge and learn to reason on object-attribute composition and dependency present in the sentence. In the downstream attribute recognition task, we propose a novel generative retrieval strategy to extract and apply the knowledge acquired from the prefixLM's large-scale pretraining. Different from contrastive retrieval, generative retrieval models the conditional dependency in a sentence, hence is more aligned with the actual language semantics. \{A\} and \{O\} are placeholders for attributes or objects in the sentence.}
    \includegraphics[width=0.8\linewidth]{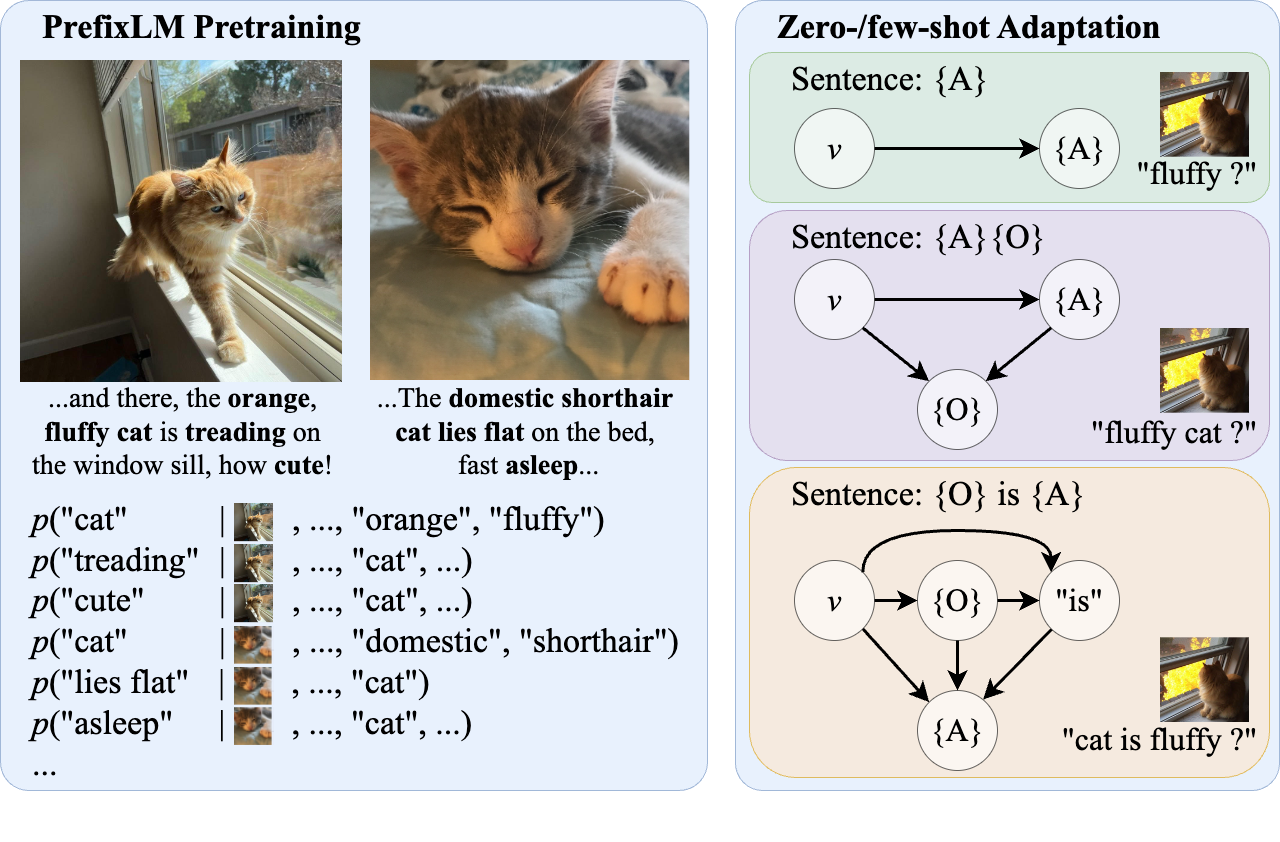}
    \label{fig:concept}
\end{figure}

However, naively applying contrastive retrieval to attribute prediction tasks is suboptimal due to its two inherent problems. First, treating input text as an unstructured whole to be aligned with images results in insufficient learning on attributes if the object alone is distinguishable enough in the image to match the image-text pair, as often is the case in CLIP training data. This creates a discrepancy between the pre-training and the downstream tasks: the model learned to primarily differentiate between objects but is later asked to understand finer attributes. Second, contrastive prompting cannot model the co-dependency between objects and attributes. Since contrastive pre-training does not capture word sequence order, as opposed to language model pre-training (Fig.~\ref{fig:concept} (left)), the model is unable to correctly measure the likelihood of counterfactual combinations such as ``bell-shaped sky'' or ``graffitied sky'' (see Fig.~\ref{fig:qualitative_examples}). These challenges emphasize the necessity for better methods in modeling object-attribute dependency in the context of large image-text foundation models.

This paper presents a novel approach to address the two aforementioned problems in applying image-text foundation models to attribute learning. The approach consists of two parts: prefix language modeling (prefixLM)~\cite{bengio2000neural,wang2021simvlm} as the pre-training foundation, and a novel, sentence generation-based formulation of attribute retrieval that allows for the extraction of pre-training knowledge for structural reasoning (see Fig.~\ref{fig:concept}).
During pre-training, the prefixLM is trained to predict the next token based on visual inputs and previous text tokens, which inherently captures diverse combinations of object-attribute dependencies in the sentence. In the downstream attribute recognition task, we measure the object-attribute alignment in an image by evaluating the probability of generating a sentence capturing the relations. We refer to this approach as ``generative retrieval''. In particular, this method enables flexible retrieval for a wide range of attribute relations (associative, possessive, or further modified with temporal words like "currently" or "already") through building arbitrary conditional dependency models for downstream tasks at inference time (Fig.~\ref{fig:concept} (right)), which are effectively "meta-models" .

There are two immediate applications for the proposed prefixLM + generative retrieval framework: (1) describing objects through their visual appearance, state of being, or relationship to other objects in the image. And conversely, (2) recognizing objects based on their various visual attributes such as color, shape, size, and so on. In addition, our method can be further applied towards many other visual tasks that require structural reasoning. We summarize the contributions as follow:

\begin{enumerate}[nolistsep,noitemsep,leftmargin=*]
    \item We formally \textbf{reframe the attribute recognition problem} as a task of learning and modeling the image-object-attribute conditional probabilities in a large visual-language modeling setting.
    \item We establish the effectiveness of using prefixLM as a foundational model for capturing complex object-attribute relationships in \textbf{pretraining}, and propose the generative retrieval method to flexibly \textbf{distill pretraining knowledge for downstream attribute recognition tasks}.
    \item We demonstrate the limitations of purely using contrastive learning for attribute recognition and show the superior zero-shot and finetuning performance of our method.
    \item We introduce Visual Genome Attribute Ranking (\href{https://github.com/google-research/google-research/tree/master/attribute_with_prefixlm}{VGARank}), a novel benchmark combining attribute and object recognition tasks into an unified, and therefore directly comparable setting, to demonstrate the generalizability of the proposed approach.

\end{enumerate}
\section{Related Work}
\label{sec:related}

We first introduce studies on attribute learning, which mostly rely on hand-crafted probabilistic models without the use of large language models.
Then, we summarize existing works on language modeling to specifically introduce PrefixLM to the attribute learning tasks. 
Finally, we provide an overview of existing prompt learning techniques to introduce our novel approach of generative retrieval, which distillate information from the pretrained PrefixLM. 

\textbf{Visual attribute recognition} involves identifying the properties of visual objects, such as their color, material or shape. Early works had focused on object description (\textit{img}$\rightarrow$\textit{att}) using  classification~\cite{farhadi2009describing,ferrari2007learning,patterson2016coco} or relative ranking models~\cite{parikh2011relative,kovashka2012whittlesearch,chen2018compare,Wang_2016_CVPR,yu2014fine} to learn attribute presence independent of object category. Some works use attributes as a foundation~\cite{lampert2009learning,jayaraman2014zero,al2016recovering} for zero-shot object recognition (\textit{img,att}$\rightarrow$\textit{obj}; e.g., recognizing ``zebra'' by the attributes ``black'', ``white'', ``striped''). These works learn an attribute vector space and use it to recognize unseen visual objects based on the marginal probability.
In visual object detection, some models~\cite{Zhang_2021_CVPR,Anderson_2018_CVPR} train additional attribute prediction branches using the Viusal Genome dataset~\cite{krishna2017visual} to improve model diversity and to create models with multi-tasking capabilities. These models concatenate the visual feature with the ground-truth object class embedding and feed them into an attribute prediction branch (\textit{img,obj}$\rightarrow$\text{att}).

Vector space-based approaches has also been studied for attribute recognition. For example, \cite{Pham_2022_ECCV,gu2021open,chen2023ovarnet} apply the CLIP~\cite{radford2021learning}. They use the CLIP embedding to compare visual information against predefined attribute prompts (\textit{img}$\leftrightarrow$\textit{obj,att}), to determine if the image contains those attributes. In addition to CLIP,  \cite{nagarajan2018attributes,Naeem_2021_CVPR,Misra_2017_CVPR,nan2019recognizing} allow objects and attributes to be projected into the same feature space, while the decoration of attributes on objects is modeled as an operator (\textit{img}$\leftrightarrow$\textit{obj} \text{OP} \textit{att}, operator OP could be $\pm$ or linear transform).

Our innovation lies in the novel view of treating attribute recognition as a language modeling problem. We integrate probability modeling for image, object class, and attribute prediction in an unified image-text model, while leveraging LLM's foundational pre-training.

\textbf{Language modeling} (LM) estimates the probability of a sequence of words being observed in a sentence. Early language models use dense neural networks~\cite{bengio2000neural} or recursive neural networks~\cite{xu2015show}, while recent large language models (LLMs) \cite{devlin-etal-2019-bert,yu2022coca,wang2021simvlm,radford2021learning,chowdhery2022palm} are based on the transformer architecture~\cite{vaswani2017attention} because of its strong generalizability. LM has many applications in both NLP and computer vision, including question answering(QA)~\cite{rajpurkar2016squad,yang2015wikiqa}, conversational question answering~\cite{reddy2019coqa}, visual captioning~\cite{young-etal-2014-image,chen2015microsoft,sharma-etal-2018-conceptual,Agrawal_2019_ICCV}, and visual question answering ~\cite{VQA,balanced_binary_vqa,balanced_vqa_v2}. These applications can be categorized into three main types of LM: (1) image-text matching~\cite{frome2013devise}, (2) masked language modeling~\cite{devlin-etal-2019-bert}, and (3) prefix language modeling~\cite{bengio2000neural}.

Attribute recognition is a condensed VQA problem that requires predicting the visual attribute of an query object. The foundational methods proposed in the VQA domain mostly combine image-text matching and masked language modeling. Examples include LXMERT~\cite{tan2019lxmert}, UNITER~\cite{chen2019uniter}, OSCAR~\cite{li2020oscar}, VinVL~\cite{Zhang_2021_CVPR}, ViLT~\cite{kim2021vilt}, VLMo~\cite{bao2022vlmo}. 

Different from these works, we show that prefix language modeling (prefixLM)~\cite{bengio2000neural,yu2022coca,wang2021simvlm,chen2022pali,chen2023pali}) can approximate masked language modeling (see Sec.~\ref{sec:approach:conditional_dependense}) in the attribute tasks. With a novel prompting scheme, prefixLM exhibits even greater expressive power than MLM, making it a powerful tool for deep visual reasoning \cite{yao2021cpt,tsimpoukelli2021multimodal}.

\textbf{Prompt learning} originates in the field of natural language processing (NLP), where tasks like question-answering are frequently formulated as a "fill-in-the-blank" problem. Notable examples include BERT~\cite{devlin-etal-2019-bert} which employs masked language modeling, and GPT~\cite{radford2019language} that uses prefix language modeling. While large language models  (LLMs)~\cite{chowdhery2022palm,thoppilan2022lamda,openai2023gpt4} have been widely explored in NLP for fact-based reasoning, their application in the computer vision domain is relatively unexplored.

Prompt learning in computer vision has gained attention following the success of CLIP~\cite{radford2021learning}. Numerous works~\cite{Rao_2022_CVPR,yao2021cpt,Zhong_2022_CVPR,Wang_2022_CVPR,Ma_2022_CVPR,Shi_2022_CVPR,Materzynska_2022_CVPR,Pham_2022_ECCV} have focused on designing CLIP-prompts or utilizing the pre-trained CLIP checkpoint. 
Approaches such as \cite{Zhou_2022_CVPR,zhou2022learning} learn the prompting vectors instead of manually designing text prompts.

Our approach focus on its application towards attribute learning, which the aforementioned contrastive learning based methods are ill-suited for. Our proposed generative prompting is based on image-conditioned prefix language modeling~\cite{yu2022coca,wang2021simvlm}, which takes sequence ordering into consideration and is therefore well-suited for modeling the dependence between visual objects and attributes. The proposed method has potential applications in other visual reasoning problems such as visual relation detection~\cite{lu2016visual} or scene graph generation~\cite{Johnson_2015_CVPR}.

\section{Approach}
\label{sec:approach}

\subsection{Image-Conditioned Language Modeling}
\label{sec:approach:language_modeling}

Our proposed generative retrieval is based on image-conditioned prefix language modeling, i.e. image captioning. Given an image $v$, we aim to generate the corresponding text $x=(s_1, ..., s_n)$ by modeling the probability $p(x|v)$ using Eq.~\ref{eq:prefixlm}. This equation factors $p(x|v)$ into the product of conditional probabilities~\cite{bengio2000neural,radford2019language}, where at each time step, the model predicts the next token $s_i$ based on the visual input $v$ and previous tokens $(s_0, ...,s_{i-1})$ ($s_0$ is the start-of-sentence token ``\textless s\textgreater'').
\begin{equation} \label{eq:prefixlm}
\begin{split}
    p(x|v) = \prod_{i=1}^n p(s_i|v, s_1,\dots,s_{i-1})
\end{split}
\end{equation}

The factorization provided by Eq.\ref{eq:prefixlm} is advantageous as it breaks down the word generation process into individual probability factors. In Fig.~\ref{fig:concept} (left), we show that the model can capture various  object-attribute compositions during pre-training.  As a result, in downstream attribute-related tasks, we can leverage this factorization to address reasoning questions such as $p(w_{att}|v, w_{obj})$, which represents the probability of observing an attribute $w_{att}$ (e.g., ``orange'', ``fluffy'') given the visual input $v$ and object $w_{obj}$ (e.g., a ``cat'').

\subsection{Generative Retrieval for Attribute Classification}
\label{sec:approach:generative_prompting}

We formalize the simplest attribute classification task as a common foundation for both generative retrieval and contrastive retrieval. Specifically, given an image $v$ and sentence $t^{(1)},\dots,t^{(C)}$ ($C$ is number of classes), retrieval-based classification involves designing a loss function $L(v, t)$ to measure the cost of aligning image $v$ and text $t^{(i)}$ ($1\le i \le C$). Thus, zero-shot classification can be achieved through finding the class label $c = \operatornamewithlimits{argmin}_{1 \le i \le C} \{L(v, t^{(i)})\}$.

\textbf{Contrastive retrieval} builds on the fact that paired image-text are projected into the same feature space through contrastive learning during pre-training. Assuming the image is encoded as $f(v)$ and the text is encoded as $g(t)$, the contrastive learning objective aims to maximize the inner product between the matched image-text embeddings while minimizing the unmatched ones.  This encourages paired image-text samples to have a high similarity while pushing unpaired samples apart. Under the common assumption of unit norm in the embeddings \cite{sohn2016improved, jia2021scaling,radford2021learning}, this can be equivalently represented by using the L2 loss to measure the distance between image and text, denoted as $L^{(con)}(v, t) = \lVert f(v) - g(t) \rVert_2$.

\textbf{Generative retrieval} is our proposed approach for visual attribute recognition, which utilizes cross-entropy to evaluate the image-text alignment loss, represented as $L^{(gen)}(v, t) = - \sum_{i=1}^{N} \hat{p}(t_i)\log q_{\theta}(v, t_{j|j<i})$. Here, $\hat{p}(t_i) \in \mathbb{R}^{1\times V}$ ($1\le i \le N$, $N$ is the length) represents the one-hot representation of the $i$-th token of sentence $t$. To generate the information at the $i$-th step, the model $q_{\theta}$ relies on the image $v$ and all previous text tokens $t_{j|j<i}$ to produce a probability distribution $q_{\theta}(v, t_{j|j<i}) \in \mathbb{R}^{V\times 1}$ over the  vocabulary $V$. The term $-\hat{p}(t_i)\log q_{\theta}(v, t_{j|j<i}) \in \mathbb{R}^1$ represents the cross-entropy between the $i$-th token in the sentence $t$ and the model's prediction at the $i$-th step. Fig.~\ref{fig:coca} (middle) provides a visual representation of this equation.

\subsection{Modeling the Conditional Dependence}
\label{sec:approach:conditional_dependense}

\begin{figure}[t]
    \centering
    \caption{Conditional dependencies modeled by different sentence templates. Attribute recognition is modeled as a fill-in-the-blank problem for the highlighted \template{\hl{\{A\}}} in the graph. Our proposed method optimizes or approximates the joint probability of observing these graph meta-modals, all while only relying on the prefixLM pre-training.}
    \includegraphics[width=0.9\linewidth]{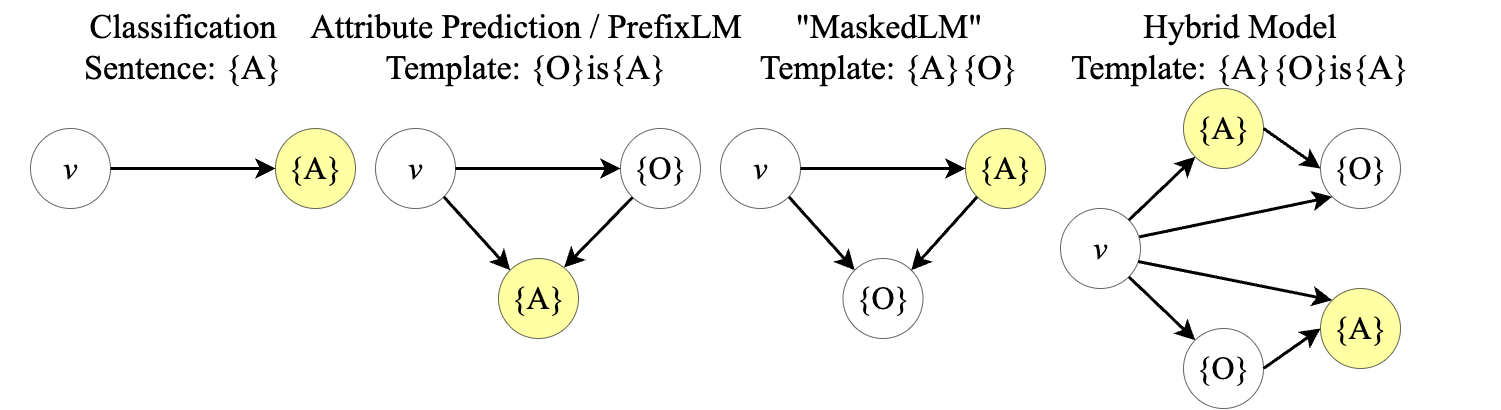}
    \label{fig:conditional_dependence}
\end{figure}

In generative retrieval, we can build different probabilistic models for visual attribute recognition by changing word ordering in the to-be-measured sentence (see Fig.~\ref{fig:conditional_dependence}).
Our key contribution to the community is proposing and showcasing its versatility in enabling the design of arbitrary probabilistic model by engineering different structure for the measured sentence. Below, we use \textit{\{A\}} to indicate attribute, and \textit{\{O\}} to indicate object.

\textbf{Sentence \template{\{A\}}.} This sentence models the simplest dependency for predicting attribute based on the image. In this dependency model, we focus on the cross-entropy of classifying the image as having a specific attribute, which can be achieved through a simple classification model. This approach aligns with early methods~\cite{farhadi2009describing,ferrari2007learning,patterson2016coco,parikh2011relative,kovashka2012whittlesearch,chen2018compare,Wang_2016_CVPR,yu2014fine} that describe attributes rather than naming the objects.

\textbf{Sentence \template{\{O\}~is~\{A\}}.} This sentence template models the prediction of attributes based on both an image and an object, approximating $p(\text{\template{\{A\}}}~|~v, \text{\template{\{O\}}})$. In this dependency model, all sentences share the same prefix  \template{\{O\}~is} (e.g., ``cat is orange'', ``cat is fluffy'', ``cat is cute'', etc.). Therefore, the only factor that matters in generative retrieval becomes $-\hat{p}(\text{\template{\{A\}}})q_{\theta}(v, \text{\template{\{O\}}}, \text{\template{is}})$, which quantifies the loss associated with classifying an attribute given the image and object. This dependency model characterizes recent attribute works such as  \cite{Zhang_2021_CVPR,Anderson_2018_CVPR,Pham_2021_CVPR,Pham_2022_ECCV}.

\textbf{Sentence \template{\{A\}\{O\}}.} This sentence is similar to Masked Language Modeling (MLM)~\cite{devlin-etal-2019-bert} as it involves filling in the blank in a sentence like ``an image of a [MASK] cat''. However, there are two key distinctions: (1) $p(\text{\template{\{A\}}}~|~v)$ requires the attribute must be easily recognizable from the image, and (2) $p(\text{\template{\{O\}}}~|~v, \text{\template{\{A\}}})$ requires that the attribute can be employed to modify the object. In contrast, MLM uses all contextual information to predict the masked token (attribute), regressing to the earlier sentence \template{\{O\} is \{A\}}). The probabilistic modeling derived from the sentence \template{\{A\}\{O\}} closely resembles the approaches in \cite{lampert2009learning,jayaraman2014zero,al2016recovering}, where attributes were utilized for object recognition.

\textbf{Sentence \template{\{A\}\{O\}~is~\{A\}}.} This sentence produces unconventional sentences such as ``fluffy cat is fluffy''. We highlight this sentence template to showcase the versatility of generative retrieval. In essence, this likelihood formulation includes all three previously discussed conditional probability terms: (1) $p(\text{\template{\{A\}}}~|~v)$ -- classification; (2) $p(\text{\template{\{O\}}}~|~v, \text{\template{\{A\}})}$ -- object-attribute compatibility; and (3) $p(\text{\template{\{A\}}}~|~v, \text{\template{\{O\}}})$ -- attribute prediction based on image and object. We present an approximate probability graph in Fig.~\ref{fig:conditional_dependence} (right), where we duplicate both the attribute and object nodes. With the duplicated \template{\{A\}} in the sentence, the resulting modeling accounts for the co-dependency between object and attribute. For example, attributes preceding objects: ``red car'', ``blue sky''; and objects preceding attributes: ``kid is smiling'', ``cat is lying''. This construction further bridges the gap between pre-training and zero-shot inference.

\textbf{Discussion.} Our proposed generative retrieval is novel for two reasons. Firstly, from the language modeling perspective, we offer a new solution for training using prefixLM, enabling the model to mimic MLM or more advanced LM in a zero-shot manner for downstream tasks. Secondly, our generative retrieval produce dependency models at inference time that serves as meta-models for attribute recognition, since we can flexibly modify the probabilistic modeling and conditional dependence through changes in sentence templates. In our experiments, we show the results for the four different probabilistic attribute models ( Fig.~\ref{fig:conditional_dependence}).

\subsection{Finetuning on Attribute Tasks}
\label{sec:approach:finetune}

Since the attribute class names have similar lengths, their cross-entropy scores $L^{(gen)}(v, t)$ with the image are expected to fall within a similar range of values (see Fig.~\ref{fig:qualitative_examples}). Therefore, an intuitive way to adapt the knowledge in a few-shot manner is to learn to ``rescale'' the retrieval scores to adapt to the new dataset priors during finetuning.
Specifically, if $t^{(c)}$ is the sentence for class $c$, we introduce learnable parameters bias $\mu_c$ and scaling factor $\sigma_c$ to adjust the $L^{(gen)}(v, t^{(c)})$, resulting in a transformed probability $p_c = \text{sigmoid}\big(- \frac{L^{(gen)}(v, t^{(c)}) - \mu_c}{\sigma_c} \big)$. This probability  $p_c$ represents the likelihood of the image-object pair being associated with attribute $c$. During finetuning, $p_c$ can be optimized using cross-entropy loss. 
In Sec.~\ref{sec:results:vaw}, we also provide the baseline results of finetuning the contrastive retrieval, using the same approach (but with loss score $L^{(con)}$ instead of $L^{(gen)}$).

For the additional parameters $\mu_c$ and $\sigma_c$, we initialize $\mu_c$ using -15.0 and $\sigma_c$ using 0.5, inspired by the values we observed in Fig.~4 (which shows $-L^{(gen)}(v, t)$ for sorting purpose). The initial values roughly transform the logits $p_c$ to zero mean and scale the standard deviation to 6.0. To finetune the model, we use a batch size of 4, a maximum text length of 16, a weight-decay of 0.01. We use the Adafactor optimizer with $\beta_1=0.9$ and $\beta_2=0.999$, and a learning rate of 1e-5 linearly decayed to zero in 100k training steps, which are roughly 1.8 training epochs. All experiments are conducted on single machine with TPUv3 with the average time to finetune a model being 7 hours.

\section{Experiments}
\label{sec:results}

\begin{figure}[t]
    \centering
    \caption{Overview of Coca. CoCa integrates both contrastive learning and prefix language modeling. While its text decoder as a whole (Unimodal+Multimodal) learns to caption images, the first few layers (Unimodal) can be used for contrastive learning.}
    \includegraphics[width=1\linewidth]{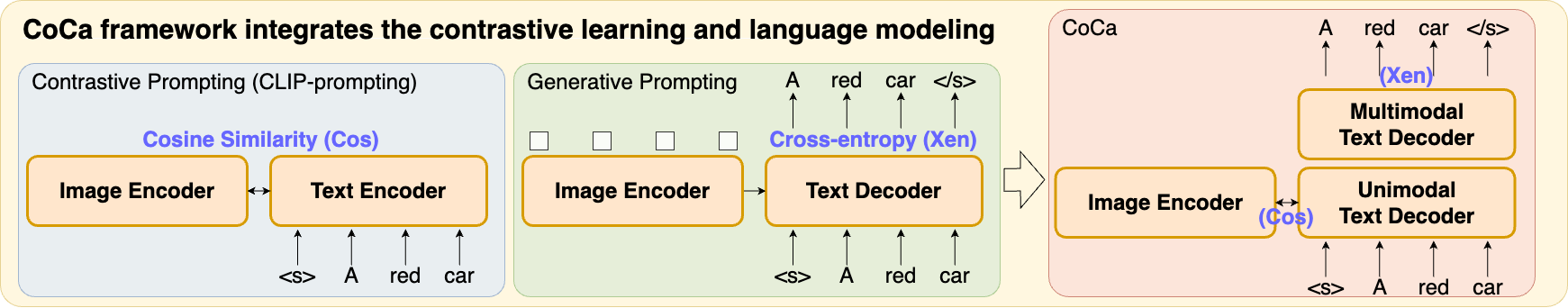}
    \label{fig:coca} 
\end{figure}

\subsection{Implementation Details}
\label{sec:results:impl}

We use CoCa~\cite{yu2022coca} pretrained on the LAION~\cite{schuhmann2022laion} as the foundation model. CoCa combines multimodal contrastive learning with image-conditioned prefix language modeling, as illustrated in  Fig.~\ref{fig:coca}. Its text decoder consists of (1) a unimodal text decoder trained on a contrastive learning objective with an image encoder, and (2) a multimodal text decoder trained to generate image captions by cross-attending to the image encoder. We adopt CoCa as the foundation model as it allows for performing both contrastive and generative retrieval with one model trained on the same image-text data, ensuring a fair comparison.
In our experiments, we use the CoCa Base model, which consists of a ViT~\cite{dosovitskiy2020image} image encoder with 12 transformer layers, a unimodal text decoder with 6 layers, and a multimodal text decoder with an additional 6 layers.  The image resolution is set to $224\times 224$ pixels with a patch size of $16\times 16$ pixels. All transformer layers have hidden dimensions of 768 and MLP size of 3,072. 

The following two datasets are used for evaluation:


\textbf{Visual Attribute in the Wild (VAW)}~\cite{Pham_2021_CVPR} is a large dataset of images with explicitly labeled positive and negative attributes. The associated task requires a model to predict a set of visual attributes given an object's name and the image it is on. VAW contains 216,790 objects from 58,565 images for training, 12,286 objects from 3,317 images for validation, and 31,819 objects from 10,392 images for testing. We use the test set to report results after validating the model.

\textbf{Visual Genome Attribute Ranking (VGARank)} is a modified version of the Visual Genome (VG) dataset~\cite{krishna2017visual} also designed to evaluate a model's ability to recognize visual attributes. The proposed dataset is different from VAW in that it is 1) an open-vocabulary ranking task, instead of a fixed vocabulary domain classification task, and 2) has two variants, VGARank-Attribute or VGARank-Object focusing on either attribute recognition given an object or object recognition given an attribute. This allows us to investigate how pretraining knowledge differs between attribute concepts and object concepts. 

For VGARank-Attribute, each ranking problem is formulated with respect to one anchor object, with $N$ ground truth attributes paired with that object in Visual Genome's annotations and $(50 - N)$ additional false attributes. VGARank-O mirrors this design, but is formulated with respect to an anchor attribute present on the image.
To make the problem challenging, these false pairings are selected in accordance to the dataset's conditional probability $P(object|attribute)$ or $P(attribute|object)$, i.e. for a given object, we select the attributes most likely to co-occur with the object in the Visual Genome distribution but which are not true for the given bounding box. Additionally, if any of the selected fake pairing exists on the current image as part of another bounding box instance, we would not include it in the set of $(50 - N)$ fake pairings. In the case where the given object or attribute does not appear often enough in Visual Genome and there is not enough fake pairing candidates from the conditional probability $P(object|attribute)$ or $P(attribute|object)$ to make up the set of $(50 - N)$, we would select fake object or attribute according to the dataset prior $P(object)$ or $P(attribute)$ to fill the rest of the choices. We obtain a dataset with 770,721 ranking problems for training, 7,997 for validation, and 32,299 for testing, and the dataset is available on our \href{https://github.com/google-research/google-research/tree/master/attribute_with_prefixlm}{GitHub page}.

\begin{figure}[t]
    \centering
    \caption{Zero-shot attribute prediction - qualitative results on the VAW dataset. Images are cropped using the yellow bounding boxes, and models only see the areas inside the boxes.}
    \includegraphics[width=1\linewidth]{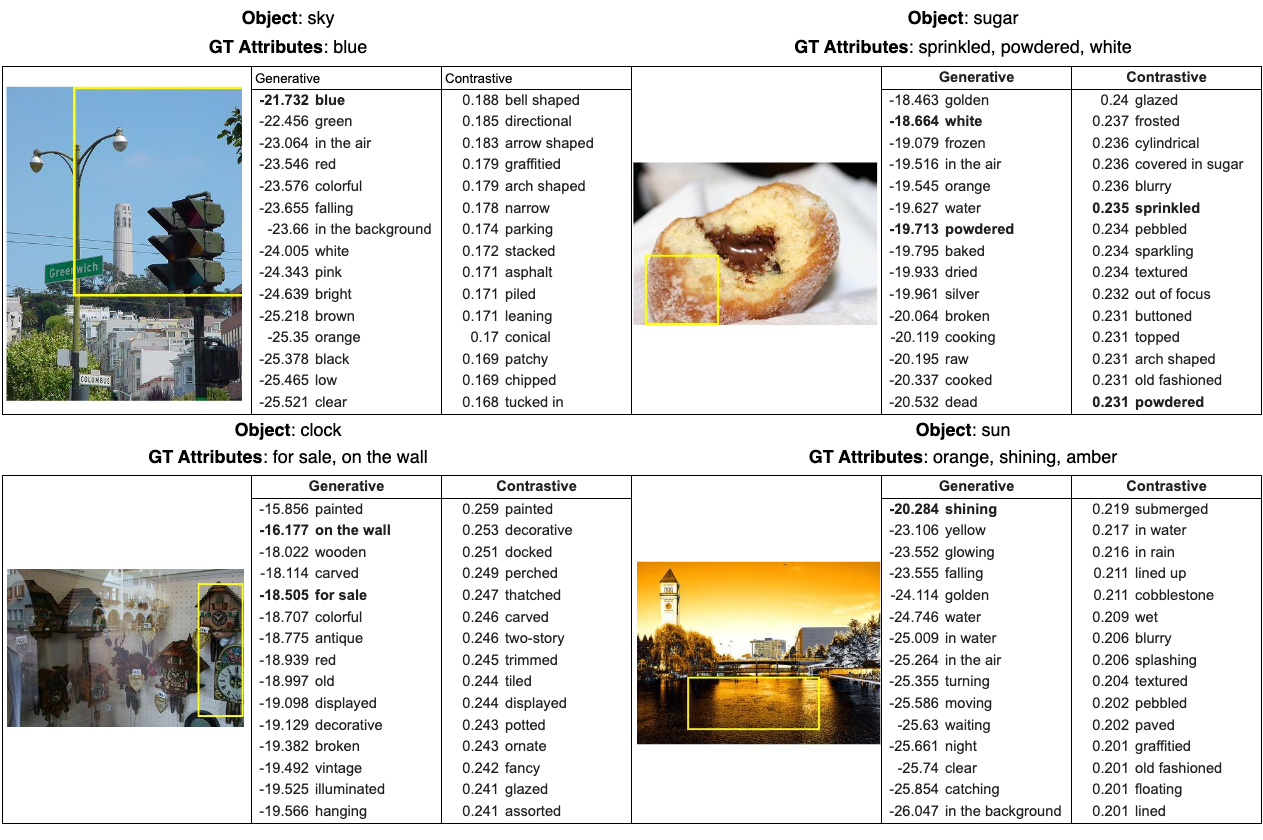}
    \label{fig:qualitative_examples}
\end{figure}

\subsection{Results on the VAW Dataset}
\label{sec:results:vaw}

First, we show that generative retrieval is better than contrastive retrieval, then analyze various conditional models, and finally compare our results to the state-of-the-art and analyze in particular its much superior performance on the less frequently seen categories.

\begin{table}
\parbox{.49\linewidth}{
    \centering
    \footnotesize
    \setlength\tabcolsep{0pt}
    \caption{Zero-shot results on VAW. Generative retrieval vs contrastive retrieval, across different sentence templates / dependency meta-models. \first{Blue} and \second{black} numbers in bold represent the best and second best, respectively. The best performing setting is highlighted in \noindent\colorbox{gray!30}{\makebox[1em]{gray}}. }
    \begin{tabularx}{1\linewidth}{lc*{3}{>{\centering\arraybackslash}X}}
        \Xhline{2\arrayrulewidth}
            & Dependency & Rank$\downarrow$ & mR$^{@15}$$\uparrow$ & mAP$\uparrow$ \\
        \hline
            \multirow{4}{*}{Con}
            &\templatee{\{A\}}                    & 95.1 & 32.0 & {52.5} \\
            &\templatee{\{A\}\{O\}}            & 149.8 & 22.4 & 47.1 \\
            &\templatee{\{O\}is\{A\}}         & 151.4 & 23.2 & 45.9 \\
            &\templatee{\{A\}\{O\}is\{A\}} & 141.0 & 23.7 & 48.3 \\
        \hline
            \multirow{4}{*}{Gen}
            & \templatee{\{A\}}                    & 82.1 & 28.7 & \first{53.8} \\
            & \templatee{\{A\}\{O\}}            & 63.9 & \first{35.9} & 47.7 \\
            & \templatee{\{O\}is\{A\}}         & \second{61.9} & \second{32.9} & 46.1  \\
            & \ourresult{\templatee{\{A\}\{O\}is\{A\}}} & \ourresult{\first{56.0}} & \ourresult{31.7} & \ourresult{\second{49.9}} \\
        \Xhline{2\arrayrulewidth}
    \end{tabularx}
    \label{tab:results:vaw_zeroshot}
}
\hfill
\parbox{.49\linewidth}{
    \centering
    \small
    \setlength\tabcolsep{0pt}
    \caption{Finetuning results on VAW. Generative retrieval vs contrastive retrieval, across different sentence templates / dependency meta-models. \first{Blue} and \second{black} numbers in bold represent the best and second best, respectively. The best performing setting is highlighted in \noindent\colorbox{gray!30}{\makebox[1em]{gray}}. }
    \begin{tabularx}{1\linewidth}{lc*{3}{>{\centering\arraybackslash}X}}
        \Xhline{2\arrayrulewidth}
            & Dependency & Rank$\downarrow$ & mR$^{@15}$$\uparrow$ & mAP$\uparrow$ \\
        \hline
            \multirow{4}{*}{Con}
            & \templatee{\{A\}}                    & 18.3 & 48.6 & 69.6 \\
            & \templatee{\{A\}\{O\}}            & 12.8 & 59.8 & 65.7\\
            & \templatee{\{O\}is\{A\}}         & 12.3 & 58.9 & 66.7 \\
            & \templatee{\{A\}\{O\}is\{A\}} & 12.2 & 59.6 & 67.3 \\
        \hline
            \multirow{4}{*}{Gen}
            & \templatee{\{A\}}                    & 18.0 & 50.5 & 71.7 \\
            & \templatee{\{A\}\{O\}}            & 11.4 & 61.8 & 70.8 \\
            & \templatee{\{O\}is\{A\}}         & \second{11.1} & \second{62.1} &  \first{72.0} \\
            & \ourresult{\templatee{\{A\}\{O\}is\{A\}}} & \ourresult{\first{10.6}} & \ourresult{\first{62.6}} &  \ourresult{\second{71.9}} \\
        \Xhline{2\arrayrulewidth}
    \end{tabularx}
    \label{tab:results:vaw_finetune}
}
\end{table}

\textbf{Generative v.s. contrastive retrieval.}
The VAW dataset and the following metrics were used: rank (average rank of the correct predictions out of all 620 choices), mR@15 (mean recall over all classes at top 15 predictions for each instance), and mAP (mean average precision over all classes). We use average rank as the primary metric as it is more direct and comprehensive at describing overall ranking performance in a large candidate space.

Tab.~\ref{tab:results:vaw_zeroshot} and \ref{tab:results:vaw_finetune} show the results of the zero-shot and fine-tuning settings, respectively. Generative retrieval outperforms contrastive retrieval in both settings, demonstrating a stronger ability to model fine-grained associations between objects and attributes. Generative retrieval achieves a rank of 56.0 with its best sentence template, compared to 95.1 ($\downarrow$ lower is better) for contrastive retrieval in the zero-shot setting (Tab.~\ref{tab:results:vaw_zeroshot}) and similarly achieves 10.6 vs 12.2 in the finetuning setting. (Tab.~\ref{tab:results:vaw_finetune}).
As previously mentioned, there are two underlying reasons for generative retrieval's superiority. First, it captures true visual attributes, while contrastive retrieval may learn superficial connections through object identities (as shown in Tab.~\ref{tab:results:vaw_zeroshot}, adding object hints in contrastive retrieval makes it perform worse). Second, it explicitly models the object-attribute relations through their dependencies and interactions, which eliminates counterfactual attribute-object pairs.
Fig.~\ref{fig:qualitative_examples} shows some qualitative examples of differences between generative retrieval and contrastive retrieval. In the top-left example, the contrastive retrieval ranks highly the attributes "sky is bell shaped" and "sky is graffitied", which are strongly associated with other objects present in the bounding box but which are not applicable to the entity of sky. This shows that contrastive retrieval can surface attributes based on image-level associations acquired from contrastive pre-training, which is highly undesirable for attribute recognition.

\textbf{Conditional dependence modeling.}
In Tab.~\ref{tab:results:vaw_zeroshot} and \ref{tab:results:vaw_finetune}, we also investigate the four types of graphical models (see Fig.~\ref{fig:conditional_dependence}) that generative retrieval approximate.
As finetuning results shows similar trends, we present the zero-shot results in Tab.~\ref{tab:results:vaw_zeroshot} (bottom).

The simple classification sentence template \template{\{A\}} does not model the important object prior and achieves the worst rank of 82.1. 
The PrefixLM sentence template \template{\{A\}\{O\}} produces a better model with a rank of 63.9, as it first classifies attributes then checks whether the attributes fit the \template{[MASK]~\{O\}}, and the MLM sentence template \template{\{O\}~is~\{A\}} has a similar rank of 61.9. We want to highlight that while all baselines on the VAW in Tab.~\ref{tab:results:vaw} are analogous to this formulation, it is not the best among the four graphical models. Therefore, improving the probability modeling in these SOTA methods can potentially further improve their performance, and our generative retrieval offers an easy way to do so.
Finally, the hybrid sentence template \template{\{A\}\{O\}~is~\{A\}}  performs the best with an average rank of 56.0. This is because it jointly considers three important factors: $p(\text{\template{\{A\}}~|~v)}$, $p(\text{\template{\{O\}}~|~v, \template{\{A\}}})$, and $p(\text{\template{\{A\}}~|~v, \template{\{O\}}})$, all captured by the proposed generative retrieval. In particular, it is difficult for MLM to capture the latter two factors simultaneously, and this shows how our proposed prefixLM + generative retrieval is a more flexible approach.

\begin{table}[t]
    \tiny
    \centering
    \setlength\tabcolsep{1.5pt}
    \footnotesize
    \caption{Comparing to the SOTA on the VAW dataset. The top rows show the baseline models; the last three rows shows the results of our method, finetuned with generative retrieval sentences. On the best baseline model, TAP, we primarily compare against the version without in-domain pretraining (LSA) on the evaluation dataset. The performance of the version of TAP with in-domain pretraining is included for completeness. For metric mA, we report mA@threshold=0.005 as cross-validated. Due to space constraints, we are moving the overall mR@15 and f1@15 to the supplementary material.}
    \begin{tabularx}{1\linewidth}{l*{2}{>{\centering\arraybackslash}X}|*{3}{>{\centering\arraybackslash}X}|*{7}{>{\centering\arraybackslash}X}}
    \Xhline{2\arrayrulewidth}
        \multirow{2}{*}{Methods}
        &\multicolumn{2}{c}{Overall}
        &\multicolumn{3}{c}{Class Imb. mAP} 
        &\multicolumn{5}{c}{Attribute Type mAP} 
        \\
        & mAP & mA
        & Head & Med. & Tail
        & Colr. & Mat. & Shap. & Size & Txtr. & Actn. & Other \\
    \Xhline{1\arrayrulewidth}
        ResNet-Bas.-CE
        & 56.4 & 50.3 & 64.6 & 52.7 & 35.9 & 54.0 & 64.6 & 55.9 & 56.9 & 54.6 & 47.5 & 59.2 \\
        LSEP
        & 61.0 & 67.1 & 69.1 & 57.3 & 40.9 & 56.1 & 67.1 & 63.1 & 61.4 & 58.7 & 50.7 & 64.9 \\
        PartialBCE+GNN
        & 62.3 & 68.9 & 70.1 & 58.7 & 40.1 & 57.7 & 66.5 & 64.1 & 65.1 & 59.3 & 54.4 & 65.9 \\
        ResNet-Bas.
        & 63.0 & 68.6 & 71.1 & 59.4 & 43.0 & 58.5 & 66.3 & 65.0 & 64.5 & 63.1 & 53.1 & 66.7 \\
        ML-GCN
        & 63.0 & 69.5 & 70.8 & 59.8 & 42.7 & 59.1 & 64.7 & 65.2 & 64.2 & 62.8 & 54.7 & 66.5 \\
        Sarafianos2018\cite{sarafianos2018deep}
        & 64.6 & 68.3 & 72.5 & 61.5 & 42.9 & 62.9 & 68.8 & 64.9 & \second{65.7} & 62.3 & 56.6 & 67.4 \\
        SCoNE
        & 68.3 & 71.5 & \first{76.5} & 64.8 & 48.0 & 70.4 & \second{75.6} & 68.3 & \first{69.4} & 68.4 & 60.7 & 69.5 \\
        TAP(w/o in-domain)
        & 65.4 & 67.2 & - & - & - & - & - & - & - & - & - & -  \\ 
        \gray{TAP(in-domain PT)}
        & \gray{73.4} & \gray{73.5} & \gray{77.6}	& \gray{72.9}	& \gray{58.8}	& \gray{71.6}	& \gray{74.5}	& \gray{71}	& \gray{73.4}	& \gray{70.4}	& \gray{67.9}	& \gray{77.3} \\
    \Xhline{1\arrayrulewidth}
        Ours \template{\{A\}\{O\}}
        & 70.8 & 73.7 & 74.0 & 71.0 & 58.2 & 73.1 & 75.0 & 70.9 & 61.8 & 72.2 & 68.8 & 70.7 \\
        Ours \template{\{O\}is\{A\}}
        & \first{72.0} & \first{74.7} & 74.9 & \second{72.0} & \first{60.6} & \second{75.2} & \first{76.0} & \first{72.6} & 62.9 & \first{72.7} & \second{69.6} & \first{72.0} \\
        Ours \template{\{A\}\{O\}is\{A\}}
        & \second{71.9} & \second{74.4} & \second{75.0} & \first{72.1} & \second{59.4} & \first{75.7} & 75.3 & \second{71.2} & 62.8 & \second{72.2} & \first{70.3} & \second{71.8} \\
    \Xhline{2\arrayrulewidth}
    \end{tabularx}
    \label{tab:results:vaw}
\end{table}

\textbf{Comparing to the SOTA.}
We compared our fine-tuned model to the state-of-the-art methods in Tab.~\ref{tab:results:vaw} using the following metrics from \cite{Pham_2021_CVPR}: mAP (mean average precision over all classes), mA (mean balanced accuracy over all classes), mR@15 (mean recall over all classes at top 15 predictions in each instance), and F1@15 (overall F1 at top 15 predictions). The table below focuses on fine-grained mAP metrics as the best comparison for retrieval performance. Results on mR@15 and F1@15 can be found in the supplemental.
The following baselines were considered: ResNet-Bas.-CE~\cite{Anderson_2018_CVPR,Jiang_2020_CVPR}, ResNet-Bas.~\cite{patterson2016coco}, LSEP~\cite{Li_2017_CVPR}, \cite{sarafianos2018deep}, PartialBCE + GNN~\cite{Durand_2019_CVPR}, ML-GCN~\cite{Chen_2019_CVPR}, SCoNE~\cite{Pham_2021_CVPR}, and TAP~\cite{Pham_2022_ECCV}. We thank the authors from \cite{Pham_2021_CVPR} for reimplementing/adapting all the baselines.

\textbf{By leveraging vision-based PrefixLM pretraining, our method ranks first among methods without in-domain pretraining on the target evaluation dataset.} Even compared to TAP with in-domain pretraining, our method ranks only slightly behind TAP in overall metric and outperforms TAP on most long-tail categories. 

Compared to SCoNE and TAP, our method focuses on cross-domain knowledge extraction instead of task-specific modules or training procedures to fit to the dataset domain. The SCoNE and TAP work both rely on object mask supervision and large custom datasets during training, incorporates specialized modules, and are trained and tested in the same domain. In particular, the main metric gain in the TAP work is through in-domain pretraining on LSA~\cite{Pham_2022_ECCV}. Our method, on the other hand, is designed to be generalist and compatible to large-scale pretraining, while at the same time incorporating flexible probabilistic modeling in the architecture. One major advantage of our method is its significantly better performance in the distribution long tail — the rarer attributes of the Medium (72.0\% mAP) and Tail (60.6\% mAP) attribute classes, as shown in table 3. While SCoNE and TAP’s has higher overall mAP and Head mAP, it does not necessarily mean they are better models overall, since these numbers are biased towards frequently observed attributes where it is easy for models to directly fit to the dataset, especially when it is in the same domain. Our model’s strong performance on the long tail demonstrates its ability infer on rarely seen attributes and indicates that it fits to the priors carried by the foundation models, instead of to the dataset distribution.

\begin{table}
\parbox{.49\linewidth}{
    \centering
    \footnotesize
    \setlength\tabcolsep{0pt}
    \caption{Zero-shot results on VGARank-Attribute. Generative retrieval vs. contrastive retrieval, across different sentence templates / dependency meta-models.}
    \begin{tabularx}{1\linewidth}{lc*{4}{>{\centering\arraybackslash}X}}
        \Xhline{2\arrayrulewidth}
            & Dependency & Rnk$\downarrow$ & R$^{@1}$$\uparrow$ & R$^{@5}$$\uparrow$ & R$^{@10}$$\uparrow$  \\
        \hline
            \multirow{4}{*}{Con}
            &\templatee{\{\swapslot{A}\}}            & 17.3 & 6.3 & 22.7 & 38.4  \\
            &\templatee{\{\swapslot{A}\}\{O\}}    & 16.1 & 9.7 & 28.9 & 43.6  \\
            &\templatee{\{O\}is\{\swapslot{A}\}} & 17.2 & 8.7 & 26.4 & 40.7  \\
            &\templatee{\{\swapslot{A}\}\{O\}is\{\swapslot{A}\}}  & 16.5 & 9.0 & 27.9 & 42.8 \\
        \hline
            \multirow{4}{*}{Gen}
            &\templatee{\{\swapslot{A}\}}            & 14.0 & 8.9 & 34.2 & 53.0  \\
            &\templatee{\{\swapslot{A}\}\{O\}}    & \second{13.0} & 13.9 & 41.6 & \second{58.6}  \\
            &\templatee{\{O\}is\{\swapslot{A}\}} & 13.1 & \second{15.2} & \second{42.3} & \second{58.6}  \\
            &\ourresult{\templatee{\{\swapslot{A}\}\{O\}is\{\swapslot{A}\}}} & \ourresult{\first{12.0}} & \ourresult{\first{17.6}} & \ourresult{\first{46.6}} & \ourresult{\first{62.2}}  \\
        \Xhline{2\arrayrulewidth}
    \end{tabularx}
    \label{tab:results:vg_zeroshot_att}
}
\hfill
\parbox{.49\linewidth}{
    \centering
    \small
    \setlength\tabcolsep{0pt}
    \caption{Zero-shot results on VGARank-Object. Generative retrieval vs. contrastive retrieval, across different sentence templates / dependency meta-models.}
    \begin{tabularx}{1\linewidth}{lc*{4}{>{\centering\arraybackslash}X}}
        \Xhline{2\arrayrulewidth}
            & Dependency & Rnk$\downarrow$ & R$^{@1}$$\uparrow$ & R$^{@5}$$\uparrow$ & R$^{@10}$$\uparrow$  \\
        \hline
            \multirow{4}{*}{Con}
            &\templatee{\{\swapslot{O}\}}             & 6.0 & 32.4 & 70.2 & 83.2 \\
            &\templatee{\{\swapslot{O}\}is\{A\}}  & 5.9 & 34.1 & 70.2 & 82.9 \\
            &\templatee{\{A\}\{\swapslot{O}\}}    & 5.9 & 35.3 & 70.9 & 83.2  \\
            &\templatee{\{A\}\{\swapslot{O}\}is\{A\}} & 6.0 & 34.7 & 70.0 & 82.5  \\
        \hline
            \multirow{4}{*}{Gen}
            &\templatee{\{\swapslot{O}\}}            & 6.1 & 31.3 & 70.3 & 83.2  \\
            &\templatee{\{\swapslot{O}\}is\{A\}} & \second{6.0} & 38.9 & \second{73.2} & \second{83.6}  \\
            &\ourresult{\templatee{\{A\}\{\swapslot{O}\}}}    & \ourresult{\first{5.8}} & \ourresult{\second{40.6}} & \ourresult{\first{74.2}} & \ourresult{\first{84.4}}  \\
            &\templatee{\{A\}\{\swapslot{O}\}is\{A\}} & 6.4 & \first{41.6} & 72.3 & 82.0  \\
        \Xhline{2\arrayrulewidth}
    \end{tabularx}
    \label{tab:results:vg_zeroshot_obj}
}
\end{table}

\subsection{Results on the VGARank Dataset}
\label{sec:results:vg}

\textbf{Generative vs Contrastive Retrieval.}
We make similar observations as on the VAW dataset, shown in Tab.~\ref{tab:results:vg_zeroshot_att} and \ref{tab:results:vg_zeroshot_obj}. We observe that generative retrieval sentence variations significantly outperformed the contrastive counterparts on both datasets. The best generative retrieval sentence template on VGARank-Attribute is \template{\{A\}\{O\} is \{A\}}, achieving a rank of 12.0, while the best one on VGARank-Object is \template{\{A\}\{O\}}, achieving a rank of 5.8. This again verifies our claim that generative retrieval is more optimal for attribute recognition than contrastive retrieval.

\textbf{Conditional dependence modeling.} Tab.~\ref{tab:results:vg_zeroshot_att} and \ref{tab:results:vg_zeroshot_obj} show the results on VGARank-A and VGARank-O. We boldface the targets to be predicted, which are \template{\{\swapslot{A}\}} for VGARank-A, and \template{\{\swapslot{O}\}} for VGARank-O.

As in VAW, the classification template \template{\{\swapslot{A}\}} or \template{\{\swapslot{O}\}} is the least effective, with a rank of 14.0 on VGARank-A and 6.1 on VGARank-O.
The PrefixLM template \template{\{\swapslot{A}\}\{O\}} or \template{\{\swapslot{O}\}~is~\{A\}} performs better with rank of 13.0 and 6.0, which is expected as it first classifies the target token then checks whether the target token fits the context.
However, the more optimally ordered MLM template \template{\{O\}~is~\{\swapslot{A}\}} or \template{\{A\}\{\swapslot{O}\}} mostly outperforms the previous approach, which aligns with conclusions drawn by earlier works like \cite{lampert2009learning,jayaraman2014zero,al2016recovering} that suggest attributes help the classification of uncommon objects, hence our model's better-than-SOTA performance on the mid and long-tail categories on VAW. Notably, \template{\{A\}\{\swapslot{O}\}} achieves the best performance on the VGARank-O. 
The Hybrid template \template{\{\swapslot{A}\}\{O\}~is~\{\swapslot{A}\}} or \template{\{A\}\{\swapslot{O}\}~is~\{A\}} performs the best on VGARank-A with a rank of 12.0, but it falls behind the \template{\{A\}\{\swapslot{O}\}} variant on VGARank-O. We attribute this to the more challenging nature of attribute recognition as compared to object recognition, where the former can benefit from the more complex dependency modeling in our method, while the latter still relies more on salient information.
The VGARank-A/VGARank-O experiments highlights the versatility of the proposed method in predicting attributes from objects and vice versa. This flexibility demonstrates the foundational nature of the prefixLM approach through generative retrieval --- by simply making changes to the sentence template, we can construct various explainable probabilistic models, expanding the possibilities for modeling complex relationships between objects and attributes.

\section{Conclusion and Broader Impact}
\label{sec:conclusion}

Our work reformulates attribute learning as a probabilistic modeling problem and a knowledge extraction process from pretraining to downstream tasks, and we in turn propose the generative retrieval method on a vision-based prefixLM foundation. By leveraging the knowledge on complex word dependencies captured by the foundation model during pretraining, our work enables the explicit modeling of various object-attribute dependencies in downstream attribute tasks. We also showcase the method's flexibility in emulating various conditional probabilistic dependencies, and we envision that vision-based prefixLM, through the proposed generative retrieval method, can serve as a universal framework to construct meta-models for representing and measuring complex logical relations.

As our method studies the visual attribute recognition problem in the context of large pretrained models, advancements in large vision-language models will directly translate to stronger performance on this domain. This work also benefits the broader community on image generative models by providing a better metric than CLIP for image-text alignment at a fine-grained, attribute level. This additional metric can lead to the creation of cleaner text-image datasets that has higher standards for caption details on object-attribute correctness, which would greatly benefit the training of generative models in the community.

%
%
\bibliographystyle{splncs04}
\bibliography{main}

\newpage

\newcounter{appendixalpha}
\renewcommand{\theappendixalpha}{\Alph{appendixalpha}}
\newcommand{\appendixsection}[1]{%
    \refstepcounter{appendixalpha}%
    \section*{\theappendixalpha. #1}%
}
\makeatother

\appendixsection{Limitations}
One limitation to our proposed method is its increased computational cost. Generative retrieval has $n$ autoregressive text decoding steps, where $n$ is the length of the retrieval template sentence, while contrastive retrieval has one text encoding step. Given the short and fixed-length sentence templates in the attribute learning context, the computational complexity of generative retrieval is $n\times$ constrastive ($n$ = 2 to 4). In addition, the text-only attribute embeddings in contrastive retrieval can be precomputed and cached in advance, which would make contrastive retrieval take 0 encoding steps at inference time. This is not possible for generative retrieval, as it is not possible to precompute a part of the likelihood of generating a image-object-attribute triple.
Another limitation to the generative retrieval approach is that is is specifically designed for tasks where the assumed lengths of answers or prompts are similar. Since the sum of log probabilities in  $L^{(gen)}$ is influenced by the length of the text, the approach is biased towards shorter answers. In the context of attribute prediction tasks, the assumption of similar lengths holds true, allowing us to treat attribute prompt optimization as joint probability optimization in a graph model. This task formulation sets it apart from VQA tasks, which typically involve multiple-choice questions with answers of varying lengths. It is worth noting that this limitation does not undermine our main contribution, which is the development of a novel formulation and framework that connects knowledge from large-scale prefixLM pre-training to the method of generative retrieval for attribute recognition problems.

\appendixsection{Additional Qualitative Examples}
We provide more examples to compare our zero-shot retrieval methods, we also include the results from the fully-supervised method SCoNE~[14] trained on the VAW dataset. Fig.~\ref{fig:supp_qualitative_examples} at the end of the supplementary material shows the results. Some interesting observations can be made. First, VAW is still a closed domain dataset, lacking in the coverage of long-tailed attributes. In example (2), our generative retrieval predicts ``decorative’’, ``antique’’, and ``bamboo'', which are visually salient and grammatically correct. However, the ground-truth annotation does not include these two options. Second, compared to others, generative retrieval can surface some of the most significant attributes in the examples. For example, ``in the background’’, ``decorative’’, ``worn’’, or ``closed’’. However, many predictions of the contrastive retrieval method are visually imperceptible or incorrect, such as arch-shaped, standing, partially-eaten, water.

\begin{figure}[ht!]
    \centering
    \includegraphics[width=1.0\linewidth]{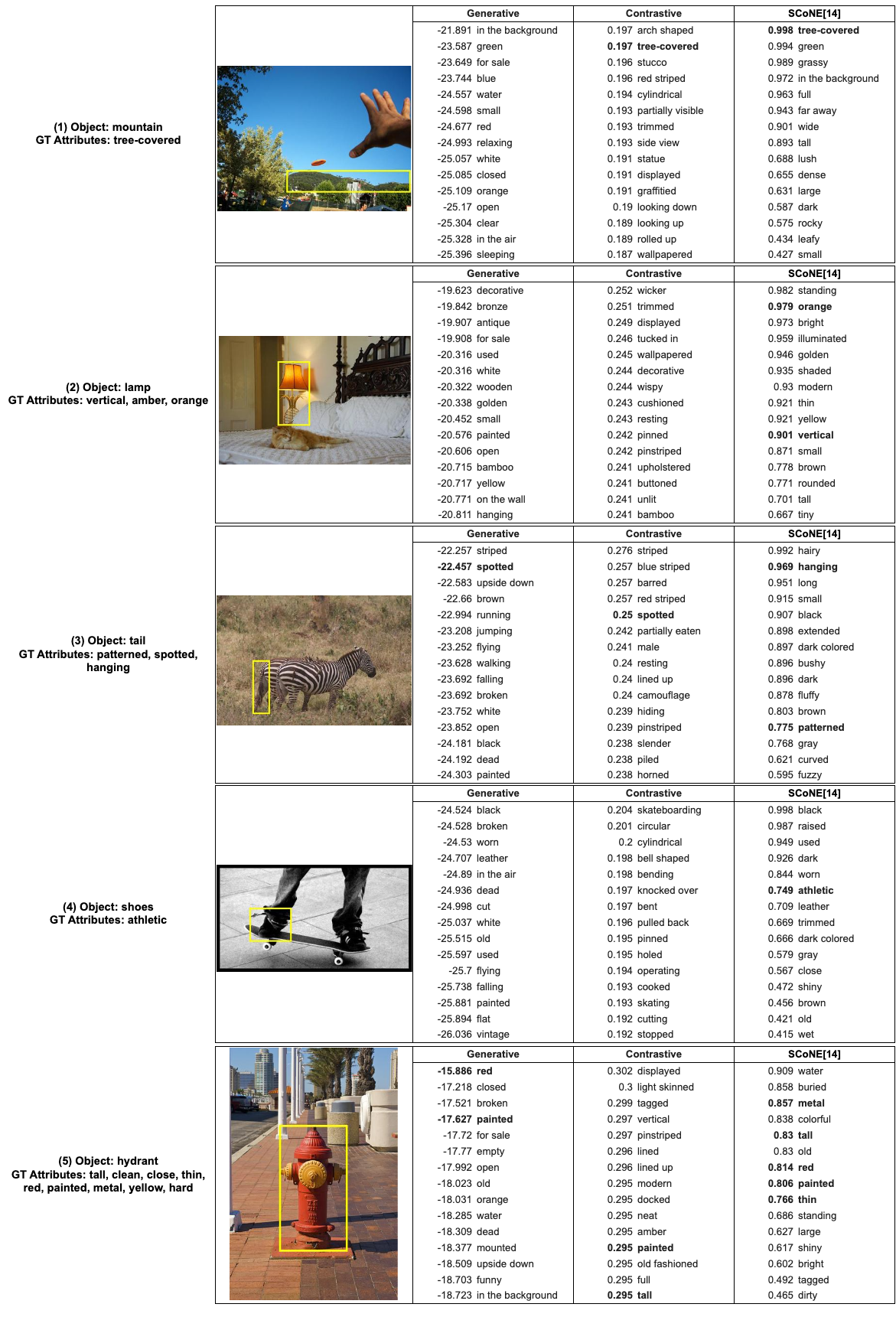}
    \caption{More qualitative examples on the VAW dataset, \textbf{zero-shot vs. fine-tuned}. The generative and contrastive columns use zero-shot retrieval, while the baseline column SCoNE [14] is fine-tuned on the VAW dataset.}
    \label{fig:supp_qualitative_examples}
\end{figure}

\appendixsection{Additional Evaluation Results}
We include additional results on the VAW experiments in Tab.~\ref{tab:results:vaw_more}, including the less comparable metrics of mR$^{@15}$ and F1$^{@15}$, which were omitted in the main text due to space constraints. Our method achieves the second place only slightly behind TAP, despite focusing more on cross-domain knowledge extraction and not on constructing task-specific models, which may involve fitting to the evaluation dataset at hand using specialized modules, training procedures, or special training data like segmentation masks that are expensive or impossible to scale.

\begin{table}[t]
    \centering
    \setlength\tabcolsep{1.5pt}
    \footnotesize
    \caption{Comparing to the SOTA on the VAW dataset. The top rows show the baseline models; the last three rows shows the results of our method which finetunes the generative prompts. For mA, we report mA@threshold=0.005 as we cross-validated.}
    \begin{tabularx}{1\linewidth}{l*{4}{>{\centering\arraybackslash}X}}
    \Xhline{2\arrayrulewidth}
        \multirow{2}{*}{Methods}
        &\multicolumn{4}{c}{Overall}\\
        & mAP & mR$^{@15}$ & mA & F1$^{@15}$ \\
    \Xhline{1\arrayrulewidth}
        ResNet-Bas.-CE
        & 56.4 & 55.8 & 50.3 & 61.5 \\
        LSEP
        & 61.0 & 50.7 & 67.1 & 62.3 \\
        PartialBCE+GNN
        & 62.3 & 52.3 & 68.9 & 63.9 \\
        ResNet-Bas.
        & 63.0 & 52.1 & 68.6 & 63.9 \\
        ML-GCN
        & 63.0 & 52.8 & 69.5 & 64.1 \\
        Sarafianos2018\cite{sarafianos2018deep}
        & 64.6 & 51.1 & 68.3 & 64.6 \\
        SCoNE
        & 68.3 & 58.3 & 71.5 & \first{70.3} \\
        TAP (w/o in-domain PT)
        & 65.4 & 54.2 & 67.2 & 66.4 \\
        \gray{TAP (in-domain PT)}
        & \gray{73.4} & \gray{63.3} & \gray{73.5} & \gray{71.1} \\ 
    \Xhline{1\arrayrulewidth}
        Ours \template{\{A\}\{O\}}
        & 70.8 & 61.8 & 73.7 & 68.3 \\
        Ours \template{\{O\}~is~\{A\}}
        & \first{72.0} & \second{62.1} & \first{74.7} & \second{68.7} \\
        Ours \template{\{A\}\{O\}~is~\{A\}}
        & \second{71.9} & \first{62.6} & \second{74.4} & \second{68.7} \\
    \Xhline{2\arrayrulewidth}
    \end{tabularx}
    \label{tab:results:vaw_more}
\end{table}

Furthurmore, to qualitatively demonstrate our model's superior performance on the less frequent categories in the distribution long tail of the Medium (72.0\% mAP vs 64.8\% mAP) and Tail (60.6\% mAP vs 48.0\% mAP) attribute classes, we show below Tab.~\ref{tab:results:qual} of model performance on the least frequent attributes in VAW:

\begin{table}[t]
    \centering
    \small
    \setlength\tabcolsep{1.5pt}
    \footnotesize
    \caption{Model performance on tail-distributed attribute categories in VAW}
    \begin{tabularx}{1\linewidth}{l*{2}{>{\centering\arraybackslash}X}}
    \Xhline{2\arrayrulewidth}
        \multirow{2}{*}{Methods}
        &\multicolumn{2}{c}{Model}\\
        & SCoNE mAP & Our mAP \\
    \Xhline{1\arrayrulewidth}
        nylon & 0.6984 & 0.5333 \\
        bell shaped & 0.6955 & \textbf{0.9167} \\
        braided     & 0.3893 & \textbf{0.7046} \\
        styrofoam   & 0.3591 & 0.3354 \\
        spiral      & 0.2294 & \textbf{0.8605} \\
        kissing     & 0.0409 & \textbf{0.4085} \\
        wallpapered & 0.5293 & \textbf{0.8956} \\
        smoking     & 0.1966 & \textbf{0.3671} \\
        stucco      & 0.3774 & \textbf{0.5914} \\
        cubed       & 0.1102 & \textbf{0.4258} \\
    \Xhline{1\arrayrulewidth}
        TAIL MEAN   & 0.4800 & \textbf{0.5940} \\
    \Xhline{2\arrayrulewidth}
    \end{tabularx}
    \label{tab:results:qual}
\end{table}

\appendixsection{Image Attribution}
In this paper we display several images from the VAW dataset. The Flickr links and the license information for these images can be found in Tab.~\ref{tab:supp_results:vaw_license}. We thank the original photographers for sharing their photos.

\begin{table}[ht]
    \centering
    \setlength\tabcolsep{2pt}
    \small
    \caption{Flickr links and license of the images.}
    \begin{tabularx}{1\linewidth}{lll}
    \Xhline{2\arrayrulewidth}
        Flickr link & User & License \\
    \Xhline{1\arrayrulewidth}
        \multicolumn{3}{l}{\textbf{Paper Fig.~4 (from left to right, top to bottom)}}\\
        \url{flickr.com/photos/mount_otz/31929683/} & mount\_otz & CC BY-NC-SA 2.0 \\
        \url{flickr.com/photos/jenny-pics/2381135314/} & jenny-pics & CC BY 2.0 \\
        \url{flickr.com/photos/worldofjan/2984166899/} & worldofjan & CC BY-NC 2.0 \\
        \url{flickr.com/photos/23909838@N02/3363471858/} & 23909838@N02 & CC BY-SA 2.0 \\
    \Xhline{1\arrayrulewidth}
        \multicolumn{3}{l}{\textbf{Supplementary materials Fig.~1 (from top to bottom)}}\\
        \url{flickr.com/photos/felipelopez/2660779383/} & felipelopez & CC BY-NC 2.0 \\
        \url{flickr.com/photos/afagen/2269170288/} & afagen & CC BY-NC-SA 2.0 \\
        \url{flickr.com/photos/nbarcet/2172355975/} & nbarcet & CC BY 2.0 \\
        \url{flickr.com/photos/dammit_jack/1523816737/} & dammit\_jack & CC BY-NC 2.0 \\
        \url{flickr.com/photos/mjhagen/4347200481/} & mjhagen & CC BY 2.0 \\
    \Xhline{2\arrayrulewidth}
    \end{tabularx}
    \label{tab:supp_results:vaw_license}
\end{table}

\end{document}